\pdfoutput=1
\documentclass[11pt]{article}

\usepackage[final]{naacl2021}

\usepackage{times}
\usepackage{latexsym}

\usepackage[T1]{fontenc}
\usepackage[utf8]{inputenc}

\usepackage{microtype}

\usepackage{makecell}
\usepackage{booktabs}
\usepackage{multirow}
\usepackage{amsmath}
\usepackage{graphicx}
\usepackage{multirow}
\usepackage{comment}
\usepackage{adjustbox}
\usepackage{subfig}
\usepackage{wrapfig}
\usepackage{bbding}

\usepackage{amssymb}
\usepackage[
  separate-uncertainty = true,
  multi-part-units = repeat
]{siunitx}

\usepackage{xcolor}

\title{
GraphVQA: Language-Guided Graph Neural Networks \\
for Scene Graph Question Answering
}

\author{Weixin Liang$^{1}$, Yanhao Jiang$^{1}$, Zixuan Liu$^{1}$ \\
  Stanford University, Stanford, CA 94305 \\
  \texttt{\{wxliang,jiangyh,zucks626\}@stanford.edu} \\
  }

\begin{document}

\maketitle

\begin{abstract}
    Images are more than a collection of objects or attributes --- they represent a web of relationships among interconnected objects. Scene Graph has emerged as a new modality for a structured graphical representation of images. Scene Graph encodes objects as nodes connected via pairwise relations as edges. To support question answering on scene graphs, we propose GraphVQA, a language-guided graph neural network framework that translates and executes a natural language question as multiple iterations of message passing among graph nodes. We explore the design space of GraphVQA framework, and discuss the trade-off of different design choices. Our experiments on GQA dataset show that GraphVQA outperforms the state-of-the-art model by a large margin (88.43\% vs. 94.78\%). 
Our code is available at \url{https://github.com/codexxxl/GraphVQA}
\end{abstract}
\footnotetext[1]{Equal Contribution. Authors listed in alphabetical order.}

\section{Introduction}

\begin{figure}
    \centering
    \includegraphics[width=1.0\linewidth]{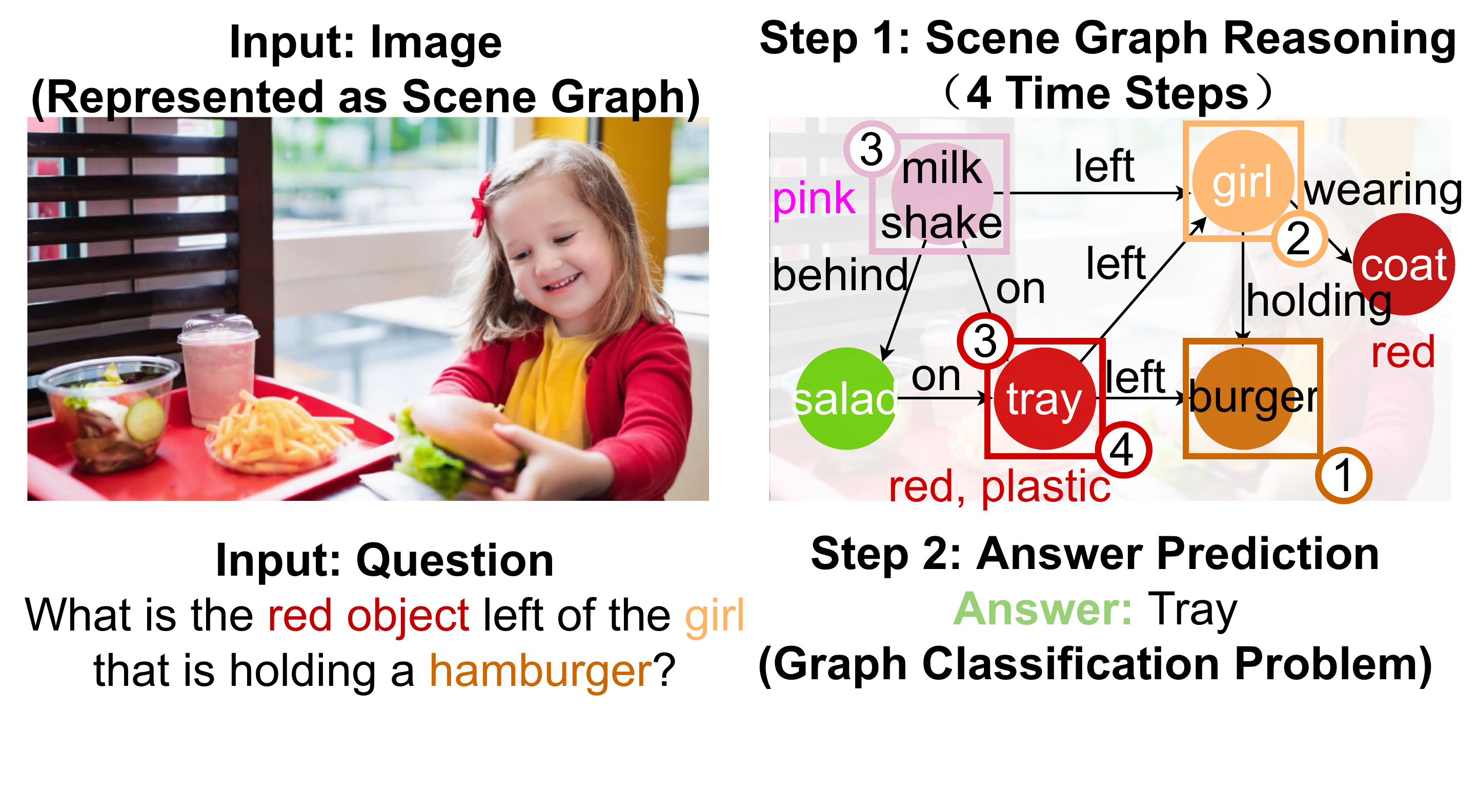}
    \vspace{-12mm}
    \caption{
    \textbf{Scene Graph:} 
    Scene graph encodes objects (e.g., girl, burger) as nodes connected via pairwise relationships (e.g., holding) as edges. 
    \textbf{GraphVQA Framework: } 
    Our core insight is to translates and executes a natural language question as multiple iterations of message passing among graph nodes (e.g., hamburger -> small girl -> red tray). 
    The final state after message passing represents the answer (e.g., tray). 
    }
    \vspace{-6mm}
    \label{fig:firstFigure}
\end{figure}

Images are more than a collection of objects or attributes. Each image represents a web of relationships among interconnected objects. 
Towards formalizing a representation for images, Visual Genome~\cite{krishnavisualgenome} defined scene graphs, a structured formal graphical representation of an image that is similar to the form widely used in knowledge base representations. 
As shown in Figure~\ref{fig:firstFigure}, scene graph encodes objects (e.g., girl, burger) as nodes connected via pairwise relationships (e.g., holding) as edges. Scene graphs have been introduced for image retrieval~\cite{ImageRetrieval}, image generation~\cite{ImageGeneration}, image captioning~\cite{ImageCaption}, understanding instructional videos~\cite{InstructionVideo}, and situational role classification~\cite{SituationRecognition}.

To support question answering on scene graphs, we propose GraphVQA, a language-guided graph neural network framework for Scene Graph Question Answering(Scene Graph QA). Our core insight is to translate a natural language question into multiple iterations of message passing among graph nodes. Figure~\ref{fig:firstFigure} shows an example question ``What is the red object left of the girl that is holding a hamburger''. 
This question can be naturally answered by the following iterations of message passing ``hamburger $\to$ small girl $\to$ red tray''. The final state after message passing represents the answer (e.g., tray), and the intermediate states reflect the model's reasoning. 
Each message passing iteration is accomplished by a graph neural network (GNN) layer. 
We explore various message passing designs in GraphVQA, and discuss the trade-off of different design choices.

Scene Graph QA is closely related to Visual Question Answering (VQA). Although there are many research efforts in scene graph generation, Scene Graph QA remains relatively under-explored. Sporadic attempts in scene graph based VQA~\cite{LCGN,DBLP:conf/iccv/LiGCL19,DBLP:conf/nips/SantoroRBMPBL17} mostly propose various attention mechanisms designed primarily for fully-connected graphs, thereby failing to model and capture the important structural information of the scene graphs.

We evaluate GraphVQA on GQA dataset~\cite{GQA}. 
We found that GraphVQA with de facto GNNs can outperform the state-of-the-art model by a large margin (88.43\% vs. 94.78\%). We discuss additional related work in appendix~\ref{sec:related}. 
Our results suggest the importance of incorporating recent advances from graph machine learning into our community.

\begin{figure*}[t]
    \centering
    \includegraphics[width=0.75\linewidth]{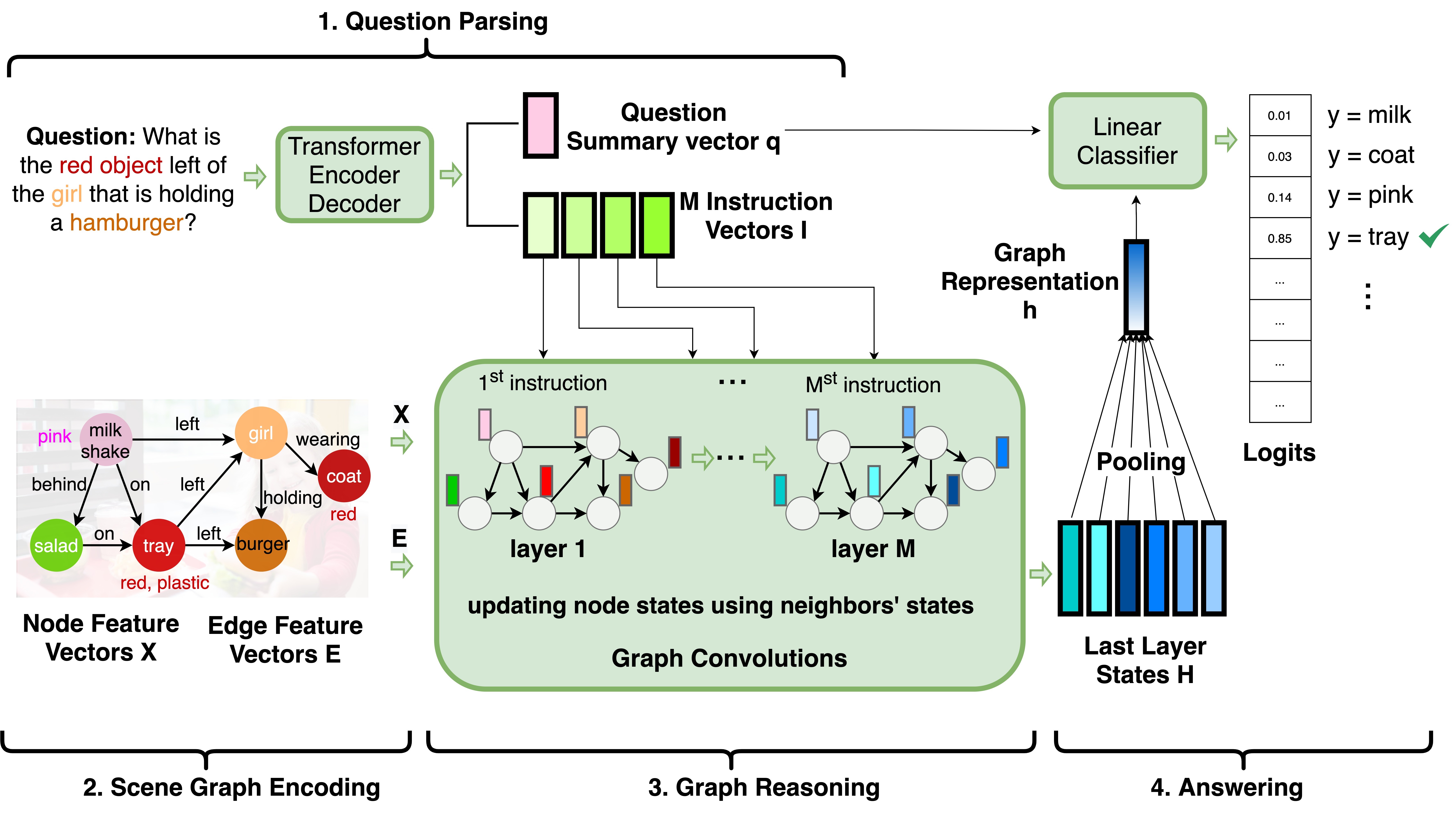}
    \caption{
    Semantics of the GraphVQA Framework. 
    (1) Question Parsing Module translates the question to $M$ instruction vectors. 
    (2) Scene Graph Encoding Module initializes node features $X$ and edge features $E$ with word embeddings. 
    (3) Graph Reasoning Module perform message passing with graph neural networks for each instruction vector. 
    (4) Answering Module summarizes the final state after message passing and predicts the answer. 
}
\vspace{-6mm}
    \label{fig:general}
\end{figure*}

\section{Machine Learning with Graphs}

Modeling graphical data has historically been challenging for the machine learning community. Traditionally, methods have relied on Laplacian regularization through label propagation, manifold regularization or learning embeddings. 
Today's de facto choice is graph neural network (GNN), which is a operator on local neighborhoods of nodes.

GNNs follow the message passing scheme. 
The high level idea is to update each node's feature using its local neighborhoods of nodes. 
Specifically, node $i$'s representation at l-th layer $\boldsymbol{h}_{i}^{(l)}$ can be calculated using previous layer's node representations $\boldsymbol{h}_{i}^{(l-1)}$ and $\boldsymbol{h}_{j}^{(l-1)}$ as: 
\begin{align}
     &\boldsymbol{h}_{\mathcal{N}_i}^{(l)} = 
     \mbox{AGG}_{j \in \mathcal{N}_i}\phi^{(l)}(\boldsymbol{h}_{i}^{(l-1)}, \boldsymbol{h}_{j}^{(l-1)}, \boldsymbol{e}_{ji})\\
    &\boldsymbol{h}_{i}^{(l)} =\gamma^{(l)} (\boldsymbol{h}_{i}^{(l-1)},  \boldsymbol{h}_{\mathcal{N}_i}^{(l)})
\end{align}
where $\boldsymbol{e}_{ji}$ denotes the feature of edge from node $j$ to node $i$, $\boldsymbol{h}_{\mathcal{N}_i}^{(l)}$ denotes aggregated neighborhood information, $\gamma^{(l)}$ and $\phi^{(l)}$ denotes differentiable functions such as MLPs, and AGG denotes aggregation functions such as mean or sum pooling.

\section{GraphVQA Framework}
\label{sec:GraphVQA_framework}
Figure \ref{fig:general} shows an overview of four modules in GraphVQA: (1) Question Parsing Module translates the question to $M$ instruction vectors. 
(2) Scene Graph Encoding Module initializes node features $X$ and edge features $E$ with word embeddings. 
(3) Graph Reasoning Module performs message passing with graph neural networks for each instruction vector. 
(4) Answering Module summarizes the final state after message passing and predicts the answer.

\subsection{Question Parsing Module}

Question Parsing Module uses a sequence-to-sequence transformer architecture to translate the question $[q_1, \dots, q_Q]$ into a sequence of instruction vectors $[\boldsymbol{i}^{(1)},  \dots, \boldsymbol{i}^{(M)}]$ with a fixed $M$. 

\begin{equation}
    [\boldsymbol{i}^{(1)}, \dots, \boldsymbol{i}^{(M)}] = \mbox{Seq2Seq}(q_1,\dots, q_Q)
\label{eq:instruction_vector}
\end{equation}

\subsection{Scene Graph Encoding Module}
Scene Graph Encoding Module first initializes node features 
$\hat{X}=[\hat{x}_1,...,\hat{x}_N]$ with the word embeddings of the object name and attributes, and edge features $E$ with the word embedding of edge type. 
We then obtain contextualized node features $X$ by: 
\begin{align}
         &\boldsymbol{x}_{i} = \sigma(
         \frac{1}{|\mathcal{N}_i|} \sum_{j \in \mathcal{N}_i}
         (W_{\mbox{enc}} \ [\boldsymbol{\hat{x}}_{j}; \boldsymbol{e_{ij}}]))
\end{align}
where $\sigma$ denotes the activation function,  $\boldsymbol{e_{ij}}$ denotes the feature of the edge that connects node $i$ and node $j$, 
and $X=[x_1,x_2,...,x_N]$ denotes the contextualized node features.

\subsection{Graph Reasoning Module}
\label{subsec:GraphReasoning}
Graph Reasoning Module is the core of GraphVQA framework. 
Graph Reasoning Module executes the $M$ instruction vectors step-by-step, with $N$ graph neural network layers. 
One major difference between our Graph Reasoning Module and standard GNN is that, we want the message passing in layer $L$ conditioned on the $L$\textsuperscript{th} instruction vector. Inspired by language model type condition~\cite{MOSS}, we adopt a general design that is compatible with \emph{any} graph neural network design: 
Before running the $L$\textsuperscript{th} GNN layer, we concatenate the $L$\textsuperscript{th} instruction vector to every node and edge feature from the previous layer. Specifically, 
\begin{align}
    & \boldsymbol{\hat{h}}_{i}^{(L-1)} = [\boldsymbol{h}_{i}^{(L-1)}; \boldsymbol{i}^{(L)}] \\
    & \boldsymbol{\hat{e}}_{ij}^{(L-1)} = [\boldsymbol{e}_{ij}^{(L-1)}; \boldsymbol{i}^{(L)}]     
\end{align}
where $\boldsymbol{\hat{h}}_{i}^{(L-1)}$ and $\boldsymbol{\hat{e}}_{ij}^{(L-1)}$ denotes the node feature and edge feature as inputs to the $L$\textsuperscript{th} GNN layer. 
Next, we introduce three standard GNNs that we have explored, starting from the simplest one.

\subsubsection{Graph Convolution Networks (GCN)}
\label{subsubsec:gcn}
GCN~\cite{gcn} treats neighborhood nodes as equally important sources of information, and simply averages the transformed features of neighborhood nodes.  
\begin{align}
         &\boldsymbol{h}_{i}^{(L)} = \sigma(
         \frac{1}{|\mathcal{N}_i|} \sum_{j \in \mathcal{N}_i}
        %  \mbox{MEAN}_{j \in \mathcal{N}_i}
         (W_{\textrm{GCN}}^{(L)} \; \boldsymbol{\hat{h}}_{j}^{(L-1)}))
\end{align}

\subsubsection{Graph Isomorphism Network (GINE)}
\label{subsubsec:gine}
GIN~\cite{GIN} is provably as powerful as the Weisfeiler-Lehman graph isomorphism test. 
GINE~\cite{GINE} augments GIN by also considering edge features during the message passing: 
\begin{align*}
\boldsymbol{h}^{(L)}_i = & \boldsymbol{\Theta} ( (1 + \epsilon)
\boldsymbol{\hat{h}}^{(L-1)}_i + \\
& \sum_{j \in \mathcal{N}(i)} \mathrm{\sigma}
( \boldsymbol{\hat{h}}^{(L-1)}_j +  \boldsymbol{\hat{e}}^{(L-1)}_{j,i} ) )
\end{align*}
where $\Theta$ denotes expressive functions such as MLPs, and $\epsilon$ is a scale factor for the emphasis of the central node.

\subsubsection{Graph Attention Network (GAT)}
Different from GIN and GINE, GAT~\cite{gat} learns to use attention mechanism to weight neighbour nodes differently. 
Intuitively, GAT fits more naturally with our Scene Graph QA task, since we want to emphasis different neighbor nodes given different instruction vectors. Specifically, the attention score $\alpha^{(L)}_{ij}$ for message passing from node $j$ to node $i$ at $L$\textsuperscript{th} layer is calculated as:
\begin{align}
\boldsymbol{\alpha}^{(L)}_{ij} = & 
\mbox{Softmax}_{\mathcal{N}_i}
( \mbox{MLP}(\boldsymbol{\hat{h}}^{(L-1)}_i, \boldsymbol{\hat{h}}^{(L-1)}_j, \boldsymbol{\hat{e}}^{(L-1)}_{ij}) )
\end{align}
where $\mbox{Softmax}_{\mathcal{N}_i}$ is a normalization to ensure that the attention scores from one node to its neighbor nodes sum to $1$. 
After calculating the attention scores, we calculate each node's new representation as a weighted average from its neighbour nodes. 
\begin{align}
& \boldsymbol{h}_i^{(L)} = 
\sigma(
\sum_{j \in \mathcal{N}_i} \boldsymbol{\alpha}^{(L)}_{ij} \; \boldsymbol{W}_{\textrm{GAT}}^{(L)} \; \boldsymbol{\hat{h}}^{(L-1)}_j
)
\end{align}
where $\sigma$ denotes the activation function. 
Similar to transformer models, we use multiple attention heads in practice. 
In addition, many modern deep learning tool-kits can be incorporated into GNNs, such as batch normalization, dropout, gating mechanism, and residual connections.

\subsection{Answering Module}
After executing the Graph Reasoning module, we obtain the final states of all graph nodes after $M$ iterations of message passing $[\boldsymbol{{h}}_{1}^{(M)}, ..., \boldsymbol{{h}}_{N}^{(M)}]$. 
We first summarize the final states after message passing, and then predict the answer token with the question summary vector $q$: 
\begin{align}
& \boldsymbol{h} = \mbox{Aggregate}([\boldsymbol{{h}}_{1}^{(M)}, \boldsymbol{{h}}_{2}^{(M)}, ..., \boldsymbol{{h}}_{N}^{(M)}])\\
& \boldsymbol{y} = \mbox{Softmax}(\mbox{MLP}(\boldsymbol{h}, \boldsymbol{q}))
\end{align}
where $\boldsymbol{y}$ is the predicted answer. 
We note that GraphVQA does not require any explicit supervision on how to solve the question step-by-step, and we only supervise on the final answer prediction.

\begin{table*}[t]
% \normalsize
\small
% \footnotesize 
% \scriptsize 
\begin{center}
\tabcolsep=0.06cm
\begin{adjustbox}{max width=\textwidth}
\begin{tabular}{rrccccccc}
\cmidrule[\heavyrulewidth]{1-9}
 & \textbf{Method} &  \textbf{Binary} & \textbf{Open}  & \textbf{Consistency} & \textbf{Validity} & \textbf{Plausibility} & \textbf{Distribution} & \textbf{Accuracy} \\ 
\cmidrule{1-9}
Baseline1 & GCN       & 86.84           &        84.63        &          90.21            &      95.51             &            94.44           &            0.13           &         85.70          \\
\cmidrule{1-9}
Baseline2 & LCGN       &        90.57         &       88.43         &          93.88            &     95.40              &        93.89               &         0.16              &          88.43         \\
\cmidrule{1-9}
Ablation1  & Only Questions 
& 61.90           &        22.69        &          68.68            &      96.39             &            87.30           &            0.17           &         41.07          \\
Ablation2  & Only Scene Graphs       & 21.86           &        17.54        &          46.98            &      36.89             &            32.63           &           7.22          &         19.63          \\
\cmidrule{1-9}
Proposed1 & GraphVQA-GCN       &        92.11         &         88.37       &       95.44               &           95.5        &           94.4            &           0.12            &       90.18            \\
Proposed2 & GraphVQA-GINE       &         92.36        &        88.56        &          94.79            &       95.44            &          94.39             &         0.13              &         90.38          \\
\bf Proposed3 & \textbf{GraphVQA-GAT}    & \textbf{96.30}  & \textbf{93.37} & \textbf{98.37}       & \textbf{95.55}    & \textbf{95.15}        & \textbf{0.07}         & \textbf{94.78}    \\ 
\cmidrule[\heavyrulewidth]{1-9}
\end{tabular}
\end{adjustbox}
\caption{
Evaluation Results on GQA. 
All numbers are in percentages. 
The lower the better for distribution. 
}
\label{tab:mainresult}
\end{center}
\vspace{-8mm}
\end{table*}

\begin{comment}
\begin{table*}[t]
\vspace{+8mm}
\small
\begin{center}
\tabcolsep=0.06cm
\begin{adjustbox}{max width=\textwidth}
\begin{tabular}{rrcccc}
\cmidrule[\heavyrulewidth]{1-6}
 & \textbf{Method} &  \textbf{Binary} & \textbf{Open}  & \textbf{Consistency} & \textbf{Accuracy} \\ 
\cmidrule{1-6}
Baseline1 & GCN       & 86.84           &        84.63        &          90.21            &         85.70          \\
\cmidrule{1-6}
Baseline2 & LCGN       &        90.57         &       88.43         &          93.88            &          88.43         \\
\cmidrule{1-6}
Ablation1  & Only Questions 
& 61.90           &        22.69        &          68.68            &         41.07          \\
Ablation2  & Only Scene Graphs       & 21.86           &        17.54        &           7.22          &         19.63          \\
\cmidrule{1-6}
Proposed1 & GraphVQA-GCN       &        92.11         &         88.37       &       95.44               &       90.18            \\
Proposed2 & GraphVQA-GINE       &         92.36        &        88.56        &          94.79            &         90.38          \\
\bf Proposed3 & \textbf{GraphVQA-GAT}    & \textbf{96.30}  & \textbf{93.37} & \textbf{98.37}       & \textbf{94.78}    \\ 
\cmidrule[\heavyrulewidth]{1-6}
\end{tabular}
\end{adjustbox}
\caption{
Evaluation Results on GQA. 
All numbers are in percentages. 
The lower the better for distribution. 
}
\label{tab:mainresult}
\end{center}
\vspace{-8mm}
\end{table*}
\end{comment}

\section{Experiments}
\label{sec:experiments}
% \textbf{Dataset} \ \ \  
\paragraph{Setup}
We evaluate our GraphVQA framework on the GQA  dataset~\cite{GQA} which contains 110K scene graphs, 1.5M questions, and over 1000 different answer tokens. 
We use the official train/validation split of GQA. Since the scene graphs of the test set are not publicly available, we use validation split as test set. 
We set the number of instructions $M=5$. 
More dataset and training details are included in Appendix~\ref{sec:Addi}.

\paragraph{Models and Metrics}
We evaluate three instantiations of GraphVQA: GraphVQA-GCN, GraphVQA-GINE, GraphVQA-GAT. We compare with the state-of-the-art model LCGN~\cite{LCGN}. We discuss LCGN in appendix~\ref{subsec:LCGN}. 
We also compare with a simple GCN without instruction vector concatenation discussed in $\S$~\ref{subsec:GraphReasoning} to study the importance of language guidance. 
We report the standard evaluation metrics defined in \citet{GQA} such as accuracy and consistency.

\paragraph{Results} The first take-away message is that GraphVQA outperforms the state-of-the-art approach LCGN, even with the simplest GraphVQA-GCN. Besides, GraphVQA-GAT outperforms LCGN by a large margin (88.43\% vs. 94.78\% accuracy), highlighting the benefits of incorporating recent advances from graph machine learning. 
The second take-away message is that conditioning on instruction vectors is important. Removing such conditioning drops performance (GCN vs. GraphVQA-GCN, 85.7\% vs. 90.18\%). The third take-away message is that attention mechanism is important for Scene Graph QA, as GraphVQA-GAT also outperforms both GraphVQA-GCN and GraphVQA-GINE by a large margin (94.78\% vs. 90.38\%), 
even though GINE is provably more expressive than GAT~\cite{GIN}. 

\begin{figure}[t]
    \centering
    \includegraphics[width=1\linewidth]{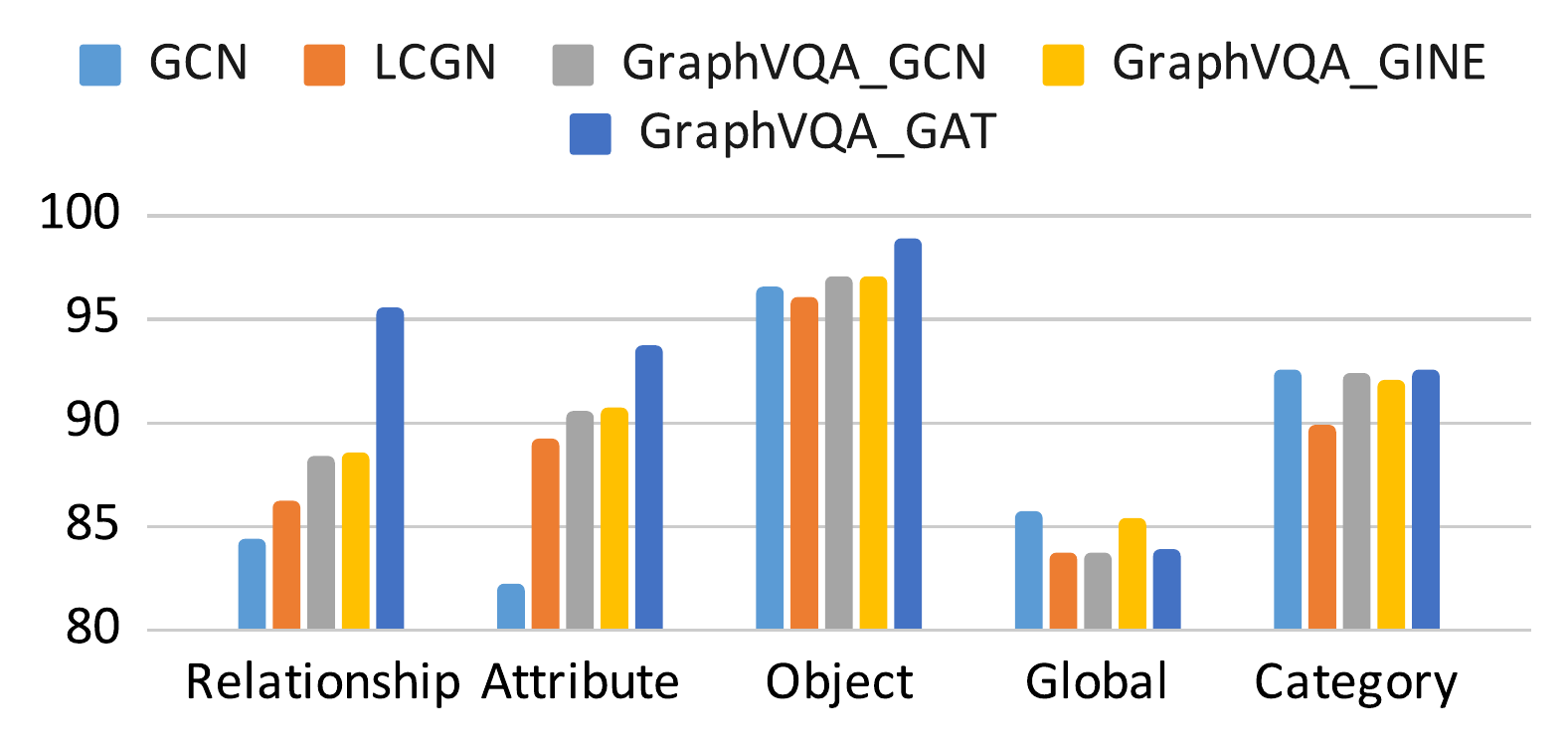}
    \caption{
    Accuracy breakdown on question semantic types. 
    GraphVQA-GAT achieves significantly higher accuracy in relationship questions (95.53\%).
}
    \label{fig:accurayTypes}
\end{figure}

\begin{figure}[t]
    \centering
    \includegraphics[width=1\linewidth]{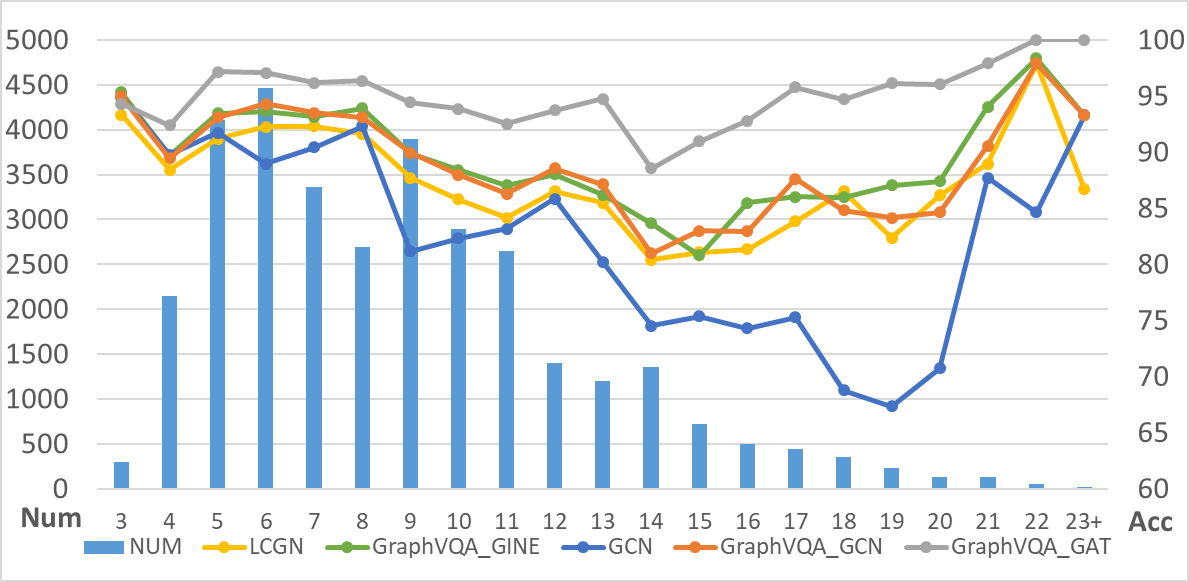}
    \caption{
    Accuracy breakdown on question word count. Num denotes the number of questions of each length. GraphVQA-GAT shows significant better performance for long question answering tasks. 
    }
    \vspace{-6mm}
    \label{fig:acc_length}
\end{figure}

\paragraph{Analysis} Figure~\ref{fig:accurayTypes} shows the accuracy breakdown on question semantic types. We found that GraphVQA-GAT achieves significantly higher accuracy in relationship questions (95.53\%). This shows the strength in the attention mechanism in modeling the relationships in scene graphs.

Figure~\ref{fig:acc_length} shows the accuracy breakdown on question word count. As expected, longer questions are harder to answer by all models. In addition, we found that as questions become longer, the accuracy GraphVQA-GAT deteriorates drops than other methods, showing that GraphVQA-GAT is better at answering long questions.

\section{Conclusion}
\label{sec:conclusion}
In this paper, we present GraphVQA to support question answering on scene graphs. GraphVQA translates and executes a natural language question as multiple iterations of message using graph neural networks. 
We explore the design space of GraphVQA framework, and found that GraphVQA-GAT (Graph Attention Network) is the best design. 
GraphVQA-GAT outperforms the state-of-the-art model by a large margin (88.43\% vs. 94.78\%). 
Our results suggest the potential benefits of revisiting existed Vision \  +  \ Language multimodal models from the perspective of graph machine learning.

\clearpage
\section*{Acknowledgments}
We would like to sincerely thank NAACL-HLT 2021 MAI-Workshop program committee for their review efforts and helpful feedback. 
We would also like to extend our gratitude to Jingjing Tian and the teaching staff of Stanford CS 224W for their valuable feedback.

\bibliography{main}
\bibliographystyle{acl_natbib}

% \clearpage
\appendix

\section{Related Work}
~\label{sec:related}

\subsection{Visual Question Answering}
VQA requires an interplay of visual perception with reasoning about the question semantics grounded in perception.
The predominant approach to visual question answering (VQA) relies on encoding the image and question with a ``black-box'' neural encoder, where each image is usually represented as a bag of object features, where each feature describes the local appearance within a bounding box detected by the object detection backbone. 
However, representing images as collections of objects fails to capture relationships which are crucial for visual question answering. 
Recent study has further demonstrated some unsettling behaviours of those models: they tend to ignore important question terms~\cite{DBLP:conf/acl/MudrakartaTSD18}, look at wrong image regions~\cite{DBLP:conf/emnlp/DasAZPB16}, or undesirably adhere to superficial or even potentially misleading statistical associations~\cite{DBLP:conf/emnlp/AgrawalBP16}. 
In addition, it has been shown that recent advances are primarily driven by perception improvements (e.g. object detection) rather than reasoning~\cite{DFOL}.

\subsection{Scene Graph Question Answering}
Although there are many research efforts in scene graph generation, using scene graphs for visual question answering remains relatively under-explored~\cite{NSM,LCGN,DBLP:conf/iccv/LiGCL19,DBLP:journals/corr/abs-2011-10731}. 
\citet{NSM} propose a task-specific graph traversal framework with neural networks. The framework requires specifying the detail ontology of the dataset (e.g., color: red, blue,...; material: wooden, metallic), and thus is not directly generalizable. 
Other attempts in graph based VQA~\cite{LCGN,DBLP:conf/iccv/LiGCL19} 
mostly explore attention mechanism on fully-connected graphs, thereby failing to capture the important structural information of the scene graphs.

\begin{figure}[t]
    \centering
    \includegraphics[width=1\linewidth]{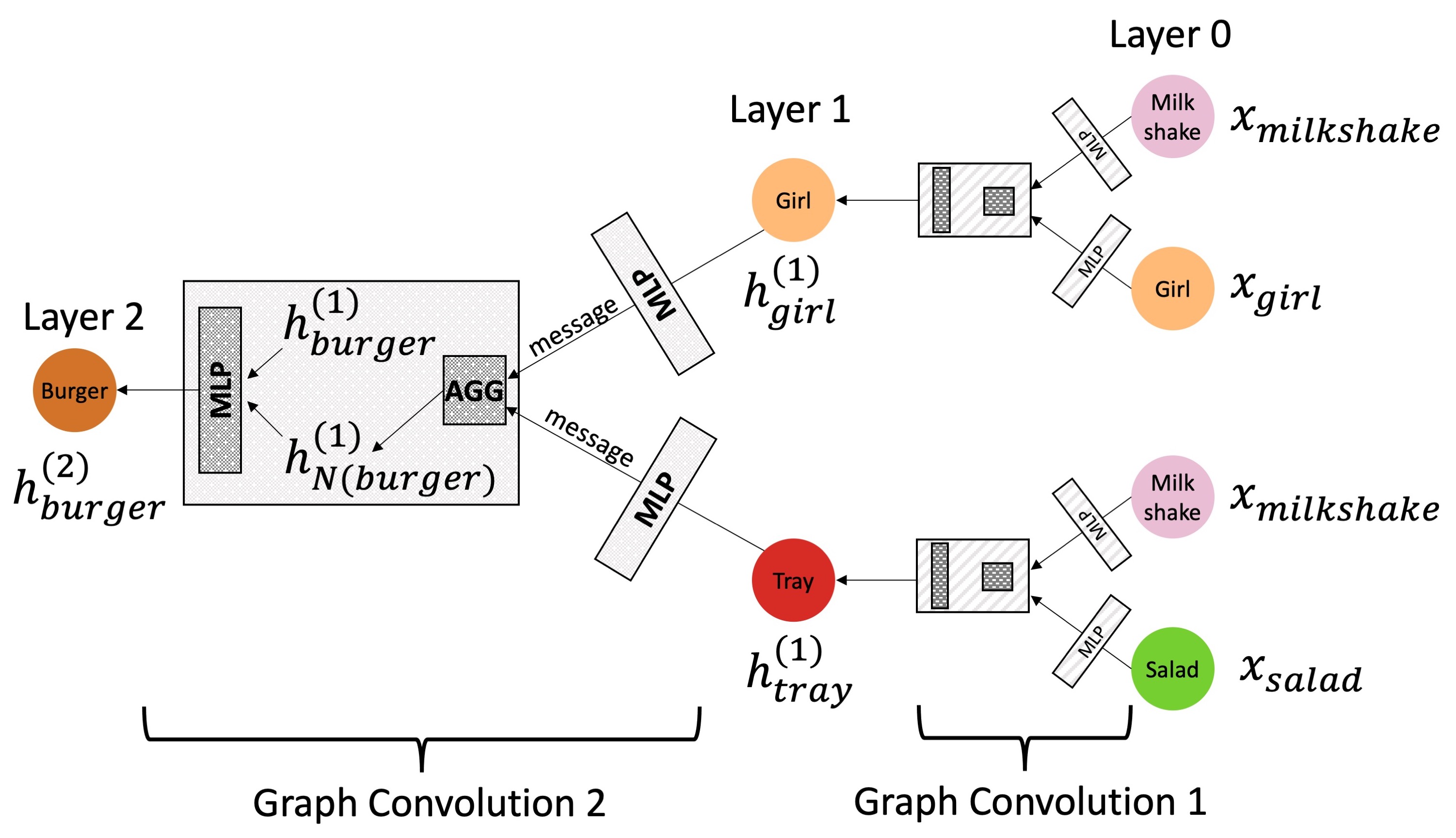}
    % \vspace{mm}
    \caption{
    \small 
    Structure of 2 Layer Graph Neural Network
}
    \label{fig:gnn}
\end{figure}

\section{Additional Results}

\subsection{Additional Performances Analysis}
\label{app_res_analysis}

\begin{figure}[t]

    \centering
    \includegraphics[width=1\linewidth]{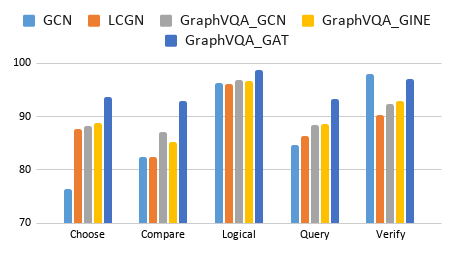}
   
    \caption{
    Accuracy breakdown on question structural types. 
    GraphVQA-GAT achieves significantly higher accuracy in all types except for verify.
} \label{fig:structural}
    
\end{figure}
Figure~\ref{fig:structural} provides another set of accuracy breakdown result on question structural types. We found that GraphVQA-GAT achieves the best for all types of questions except for the verify types.
Specifically, GraphVQA-GAT outperforms significantly than other methods on answering queries, comparing among objects and making choices. This intuitively matches the principle of attention mechanism and again shows its advantages in modeling structural information in scene graphs.

\begin{table*}[htb]
    \centering
    \small
    \begin{adjustbox}{max width=0.9\textwidth}
    \begin{tabular}[t]{rccc}
        \cmidrule[\heavyrulewidth]{1-4}

        \textbf{ Scene Graphs Statistics } & \textbf{Validation data}  & \textbf{Train data} & \textbf{All}\\
        \cmidrule{1-4}
        \textbf{Total Number of Graphs}             &      10,696 & 74,942 & 85,638   \\
        \textbf{Total Number of Nodes}                 &   174,331 & 1,231,134& 1,405,465   \\
        \textbf{Total Number of Edges}              &  534,889 & 3,795,907& 4,330,796   \\
        \textbf{Average Number of Nodes per Graph}   &  16& 16& 16   \\
        \textbf{Average Number of Edges per Graph}& 50& 51& 51   \\
        \textbf{Total Number of Node Types}           &  1,536 & 1,702& 1,703   \\
        \textbf{Total Number of Edge Types}         &  295 & 310& 310   \\
        \textbf{Total Number of Attributes Types}       &  603 & 617& 617   \\

        \cmidrule[\heavyrulewidth]{1-4}
    \end{tabular}
    \end{adjustbox}
    \caption{ 
    \small
    Scene Graphs Statistics of the GQA Dataset
    }
    \label{tab:stats}
\end{table*}

\begin{table*}[htb]
    \centering
    \begin{adjustbox}{max width=0.8\textwidth}
    \begin{tabular}[t]{llll}
        \cmidrule[\heavyrulewidth]{1-4}

        \textbf{  } & \textbf{Level of Classification}  & \textbf{Structural} & \textbf{Semantic}\\
        \cmidrule{1-4}
        \textbf{Is there apples in the picture?}              &  node & verify& object   \\
        \textbf{What color is the apple?}                 &   node & query & attribute   \\
        \textbf{Is the cat to the left or right of the flower?}             &      edge type & choose & relation   \\

        \textbf{Is it sunny or cloudy?}   &  graph & query& global   \\

        \cmidrule[\heavyrulewidth]{1-4}
    \end{tabular}
    \end{adjustbox}
    \caption{Typical types of questions
    }
    \label{tab:data}
\end{table*}
\begin{table*}[htb]
\begin{center}
\tabcolsep=0.06cm
\begin{adjustbox}{max width=0.8\textwidth}
\begin{tabular}{cccccccc}
\cmidrule[\heavyrulewidth]{1-8}
\textbf{Method} &  \textbf{Binary} & \textbf{Open}  & \textbf{Consistency} & \textbf{Validity} & \textbf{Plausibility} & \textbf{Distribution} & \textbf{Accuracy} \\ 
\cmidrule{1-8}
GraphGQA-GINE-2       & 86.83          &        83.85        &          89.8            &      95.54             &            94.25           &            0.16           &         85.04          \\

\cmidrule[\heavyrulewidth]{1-8}
\end{tabular}
\end{adjustbox}
\caption{
Ablation Study Results for 2 layer GraphVQA-GINE. All numbers are in percentages. 
The lower the better for distribution. 
}
\label{tab:ablationstudy}
\end{center}
\vspace{-5mm}
\end{table*}

\subsection{Expressive Ability Analysis of GraphVQA-GINE}
\label{comp_gcn_gine}

As mentioned in Section~\ref{subsubsec:gcn} and Section~\ref{subsubsec:gine}, a expressive function $\Theta$ is used in GINE layer. When $\Theta$ is just a single layer MLP, the corresponding GIN/GINE structure will be very similar to the GCN structure. Since in Section \ref{sec:experiments} we implemented $\Theta$ as a single layer MLP, the performance of GraphVQA-GCN and GraphVQA-GINE stays at very similar stage. As GIN and GINE are now very popular as basic components for large-scale graph neural network design, one may ask if using $\Theta$ with more powerful expression ability will help the performance. The short answer is no. We provide a simple ablation study on different choice of $\Theta$, using a two layer MLP-style network with (FC, ReLU, FC, ReLU, BN) structure. Table~\ref{tab:ablationstudy} shows that the result of GraphVQA-GINE-2 degrades to the worst. One possible reason is that the scale for each scen graph is generally small, therefore the expression ability might already be enough for a single layer MLP, and use a more complex $\Theta$ may leads to harder optimization problems, and thus leads to a downgrade of the performance. Such guess could possibly be further investigated and evaluated in our future work. 
In addition, the scene graph-based VQA as in this work might offer an opportunity for further accelerating the real world image-based applications~\cite{DBLP:journals/corr/abs-2011-10704}. Exploring such deployment benefits is another direction of future work.

\subsection{Brief Introduction of LCGN}
\label{subsec:LCGN}
Language-Conditioned Graph Networks (LCGN) \cite{LCGN} updates node representations recurrently using the same single layer graph neural network. Given a set of instruction vectors $[\boldsymbol{i}_1, \dots, \boldsymbol{i}_M]$, LCGN uses a single layer attention to convert them into context representations $[\boldsymbol{c}_1, \dots, \boldsymbol{c}_M]$. Then,
given a set of node representations $[\boldsymbol{x}_{loc,1}, \dots, \boldsymbol{x}_{loc,n}]$ , LCGN first randomly initialize another set of context representations $[\boldsymbol{x}_{ctx,1}, \dots, \boldsymbol{x}_{ctx,n}]$, and then use them to concatenate with  node representations to form  initial local features, i.e,
\begin{equation}
    \tilde{\boldsymbol{x}}_{t,i} = [\boldsymbol{x}_{loc, i}, \boldsymbol{x}_{ctx, i, t-1} , W_1\boldsymbol{x}_{loc, i} \circ W_2\boldsymbol{x}_{ctx, i, t-1}]
\label{eq:init}
\end{equation}
With the assumption that all nodes are connected, LCGN  computes the edge weights $w_{j,i}^{(t)}$for each node pair (i,j), i.e, 
\begin{equation}
    \boldsymbol{w}_{j,i}^{(t)} = \mbox{Softmax}((W_3\tilde{\boldsymbol{x}}_{t,i})^T((W_4\tilde{\boldsymbol{x}}_{t,j})\circ(W_5c_t)))
\label{eq:weight}
\end{equation}
The messages $m_{i,j}^{(t)}$, are then computed as:
\begin{equation}
    \boldsymbol{m}_{j,i}^{(t)} = \boldsymbol{w}_{j,i}^{(t)}((W_6\tilde{\boldsymbol{x}}_{t,j})\circ(W_7c_t))
\label{eq:msg}
\end{equation}
Finally, LCGN aggregates the neighborhood message information to update the context local representation $\boldsymbol{x}_{ctx, i, t}$.
\begin{equation}
    \boldsymbol{x}_{ctx, i, t} = W_8[x_{ctx,i, t-1};\sum_{j=1}^{N}m_{j,i}^{(t)}] 
\label{eq:update}
\end{equation}
Note that the graph neural structure of LCGN can be regarded as a variant of recurrently-used single standard GAT layer, but with more self-designed learnable parameters. The main difference between LCGN's and other proposed graph neural structure is that the output node and edge features will be recurrently fed into the same layer again for each reasoning step, leading to
a RNN-style network structure, instead of a sequential-style network. Moreover, our LCGN implementation is a variant of original LCGN, including a few improvements. Firstly, we use a transformer encoder and decoder to obtain instruction vectors instead of Bi-LSTM~\cite{DBLP:conf/acl/LiangZY20}. Secondly, we incorporate the true scene graph relations as edges instead of densely connected edges. Thirdly, edge attributes are also used in the generation of initial node features.

\section{Implementation Details}
\label{sec:Addi}

\subsection{Data Pre-processing}
~\label{subsec:Pre-processing}

The edges in the original scene graphs are directed. This means in most of the cases where we only have one directed edge connecting two nodes in the graph, the messages can only flow through one direction. However, this does not make sense in the natural way of human reasoning. For example, an relation of "A is to the left of B" should obviously entail an opposite relation of "B is to the right of A". Therefore, in order to enhance the connectivity of our graphs, we introduce a synthetic symmetric edge for every non-paired edge, making it pointing reversely to the source node. And in order to encode this reversed relationship, we  negate the original edge's feature vector and use it as the representation of our synthetic symmetric edge.
\subsection{Additional Dataset Information}

These scene graphs are generated from 113k images on COCO and Flicker using the Visual Genome Scene Graph~\cite{Genome} annotations. 
Specifically, each node in the GQA scene graph is representing an object, such as a person, a window, or an apple. Along with the positional information of bounding box, each object is also annotated with 1-3 different attributes. 
These attributes are the adjectives used to describe associated objects. For examples, there can be color attributes like "white", size attributes like "large", and action attributes like "standing". Attributes are important sources of information beyond the coarse-grained object classes~\cite{liang-etal-2020-alice}. 
Each edge in the scene graph denotes relation between two connected objects. These relations can be action verbs, spatial prepositions, and comparatives, such as "wearing", "below", and "taller". 

We use the official split of the GQA dataset. 
We use two files "val\_sceneGraphs.json" and "train\_sceneGraphs.json" directly obtained on the GQA website as our raw dataset. 
Since each image (graph) is independent, GQA splits the dataset by individual graphs with rough split percentages of \textbf{ train/validation: 88\%/12\%}. 
In the table \ref{tab:stats}, we summarize the statistics that we collected from the dataset. 
We did not report the statistics of the test set since the scene graphs in the test set is not publicly available.

\subsection{Training details}
We train the models using the Adam optimization method, with a learning rate of $10^{-4}$, a batch size of 256, and a learning rate drop(divide by 10) each 90 epochs. We train all models for $100$ epochs. 
Both hidden states and word embedding vectors have a dimension size of 300, the latter being initialized using GloVe~\cite{glove}.The instruction vectors have a dimension size of 512. All results reported are for a single-model settings (i.e., without ensembling). We use cross validation for hyper-parameter tuning.

\end{document}